\documentclass[conference]{IEEEtran}
\IEEEoverridecommandlockouts
\usepackage{cite}
\usepackage{amsmath,amssymb,amsfonts}
\usepackage{algorithmic}
\usepackage{graphicx}
\usepackage{textcomp}
\usepackage{xcolor}
\usepackage{ bbold }
\usepackage{ dsfont }
\usepackage{xcolor}
\usepackage{amsmath}
\usepackage{amssymb}
\usepackage{mathtools}
\usepackage{amsthm}
\def\BibTeX{{\rm B\kern-.05em{\sc i\kern-.025em b}\kern-.08em
    T\kern-.1667em\lower.7ex\hbox{E}\kern-.125emX}}
\begin{document}

\title{E-ADDA: Unsupervised Adversarial Domain Adaptation Enhanced by a New Mahalanobis Distance Loss for Smart Computing
}


\author{
\IEEEauthorblockN{Ye Gao}
\IEEEauthorblockA{\textit{Department of Computer Science} \\
\textit{University of Virginia}\\
}
\and
\IEEEauthorblockN{Brian Baucom}
\IEEEauthorblockA{\textit{Department of Psychology} \\
\textit{University of Utah}\\
}
\and
\IEEEauthorblockN{Karen Rose}
\IEEEauthorblockA{\textit{College of Nursing} \\
\textit{Ohio State University}\\
}
\and
\IEEEauthorblockN{Kristina Gordon}
\IEEEauthorblockA{\textit{Department of Psychology} \\
\textit{University of Tennessee, Knoxville}\\
}
\and
\IEEEauthorblockN{Hongning Wang}
\IEEEauthorblockA{\textit{Department of Computer Science} \\
\textit{University of Virginia}\\
}
\and
\IEEEauthorblockN{John A. Stankovic}
\IEEEauthorblockA{\textit{Department of Computer Science} \\
\textit{University of Virginia}\\
}

}

\maketitle

\begin{abstract}
In smart computing, the labels of training samples for a specific task are not always abundant. However, the labels of samples in a relevant but different dataset are available. As a result, researchers have relied on unsupervised domain adaptation to leverage the labels in a dataset (the source domain) to perform better classification in a different, unlabeled dataset (target domain). Existing non-generative adversarial solutions for UDA aim at achieving domain confusion through adversarial training. The ideal scenario is that perfect domain confusion is achieved, but this is not guaranteed to be true. To further enforce domain confusion on top of the adversarial training, we propose a novel UDA algorithm, \textit{E-ADDA}, which uses both a novel variation of the Mahalanobis distance loss and an out-of-distribution detection subroutine. The Mahalanobis distance loss minimizes the distribution-wise distance between the encoded target samples and the distribution of the source domain, thus enforcing additional domain confusion on top of adversarial training. Then, the OOD subroutine further eliminates samples on which the domain confusion is unsuccessful. We have performed extensive and comprehensive evaluations of E-ADDA in the acoustic and computer vision modalities. In the acoustic modality, E-ADDA outperforms several state-of-the-art UDA algorithms by up to 29.8\%, measured in the f1 score. In the computer vision modality, the evaluation results suggest that we achieve new state-of-the-art performance on popular UDA benchmarks such as Office-31 and Office-Home, outperforming the second best-performing algorithms by up to 17.9\%.
\end{abstract}




\section{Introduction}
Domain adaptation (DA) has drawn a lot of interest \cite{zhao2021transfer, tzeng2014deep, zhang2019manifold} because it deals with the problem that arises within a core assumption of machine learning: machine learning assumes that the testing samples are from the domain of the training samples. This assumption often results in the fact that the machine learning model's testing performance is significantly worse than its validation performance when the training and testing samples are from different distributions or domains. Unsupervised Domain Adaptation (UDA) \cite{wang2020rethink, tzeng2017adversarial} is a popular sub-field of DA because it allows the target domain to be unlabeled, which is more typical for real-life smart computing applications as a lot of samples collected from real environments are not labeled. However, if the labels of samples in a relevant, but different dataset are available, UDA can be performed to leverage the label information in a source dataset to perform better classification in a different, unlabeled dataset (target domain).

Among the methods developed for UDA, adversarial-based methods are very popular. There are two types of adversarial-based methods: generative, and non-generative. Generative methods \cite{xu2020adversarial, hoffman2018cycada}, inspired by GAN \cite{goodfellow2020generative}, aim at generating samples that aid in the task of (unsupervised) DA. For example, Hoffman et al. \cite{hoffman2018cycada} try to adapt the source samples in the style of the target domain. The resulting adapted samples can be used to train a classifier using the labels of these adapted source samples to classify the samples in the target domain. Non-generative methods \cite{tzeng2017adversarial, ganin2016domain} aims at domain confusion. It usually involves one or more generators/encoders, a domain discriminator, and a category classifier \cite{tzeng2017adversarial} that does the final classification of samples. The generator/encoder and the discriminator engage in a mini-max game in which the generator/encoder tries to deceive the domain discriminator, masking the true origin (which can be the source or target) of an incoming sample.

In this paper, we study non-generative adversarial methods for UDA and uncover a challenge still existing in the field of UDA: We observe that, while these methods attempt to maximize domain confusion via adversarial training, their effort to achieve domain confusion is implicit rather than explicit. There still exists room for improvement on the task of domain confusion if domain confusion can be attempted to be achieved both implicitly via adversarial training and explicitly. To further enforce domain confusion (explicitly) and address the challenge, we introduce a novel variation of the Mahalanobis distance loss. The original Mahalanobis distance \cite{lee2018simple} measures how one sample deviates from a distribution. The Mahalanobis distance loss is a loss function used to train the encoder/generator, which aims at making the encoder/generator achieve the minimization of the distribution-wise distance between the source samples and the encoded target samples, or vice versa.

\textbf{There are two novelties in our Mahalanobis distance loss function.} First, although the idea of Mahalanobis distance loss has been defined \cite{wen2022rotated}, it is defined by taking the true values and predicted values as input. In other words, the previously defined Mahalanobis distance loss \cite{wen2022rotated} minimizes the distance between a predicted value and the distribution of the set of true values. Our Mahalanobis distance loss, on the other hand, minimizes the distance between two distributions (the source domain distribution and the masked/encoded target domain distribution) instead of one value and one distribution, as our goal is to make the masked/encoded target domain, not an individual sample, closer to the distribution of the source domain, so that domain confusion is achieved. Second, to the best of our knowledge, we are the first to apply the Mahalanobis distance in the field of (unsupervised) non-generative adversarial domain adaption to achieve domain confusion.

We also investigate if it is possible to improve the performance of UDA tasks even further. We hypothesize that two, instead of one, category classifiers are needed. One is trained on the source samples and their (true) labels. The other is trained on the target samples and their (pseudo) labels. Then, we use an out-of-distribution (OOD) detection subroutine to determine if an encoded sample should be classified by the source category classifier or the target category classifier. The out-of-distribution is facilitated once again via the original Mahalanobis distance, as we have found studies that compare various OOD detection approaches' efficacy, and the Mahalanobis distance wins. For more details, please see Section \ref{sec:related_works}.

In addition to the Mahalanobis distance loss and the OOD detection subroutine, we use the architecture of ADDA \cite{tzeng2017adversarial} in which the source encoder (not a generator) and the source category classifier are trained end-to-end with source samples and their true labels. A target encoder (a generator) and a domain discriminator engage in a mini-max game whereas the target encoder tries to mask the incoming target samples as source-passing to fool the domain discriminator, thus achieving domain confusion. The encoded target samples are sent to the source category classifier for classification. Extensive evaluations show the superiority of the improved ADDA (we call it \textit{E-ADDA} or the Enforced ADDA), Mahalanobis distance loss-enhanced, OOD detection subroutine-enhanced, over the vanilla ADDA and various state-of-the-art algorithms. These solutions achieve significant improved state-of-the-art performance on popular UDA benchmarks such as Office-31 and Office-Home.

The contributions of this paper are:

\begin{itemize}
    \item We identify the room for improvement of existing non-generative methods for (unsupervised) domain adaptation because they solely rely on the adversarial training to achieve domain confusion, which is implicitly achieved.
    \item We introduce a new loss function that minimizes the distribution-wise distance between the source distribution and the masked/encoded target distribution to further enforce domain confusion that is experimentally superior to adversarial training alone.
    \item Our solution, E-ADDA, outperforms various state-of-the-art domain adaptation/transfer learning algorithms on the acoustic modality in the field of domain-adapting/transfer-learning from angry voices to speech of verbal conflict by up to a 29.8\% improvement in f1 scores.
    \item To further demonstrate the generalizability of E-ADDA, we also evaluate it against various state-of-the-art domain adaptation algorithms in the modality of computer vision. E-ADDA outperforms the state-of-the-art algorithms by up to 17.9\% improvement in accuracy scores on popular UDA benchmarks such as Office-31 and Office-Home.
\end{itemize}

\section{Related Work}
\label{sec:related_works}
 
 \subsection{Unsupervised Domain Adaptation}
 
  
  A large amount of work has been done on UDA by minimizing the dissimilarity between the distributions of the source and target domains. The common measurements of domain dis-similarity include KL divergence, and maximum mean discrepancy (MMD). Extensive research on transfer learning is dedicated to minimizing the dis-similarity measurements \cite{zhao2021transfer}. The minimization of dis-similarity measurements is also used with other measurements, such as classification loss on the source to find features that both discriminate and are domain-invariant \cite{tzeng2014deep}. However, the minimization of MMD of domains jeopardizes the locality structure of samples and potentially reduces the effectiveness of transfer learning \cite{zhang2019manifold}. Also, feature discriminability is also decreased due to the unintentional minimization of joint variance of features from source and target sets \cite{wang2020rethink}. 

  Adversarial-based UDA has been a popular sub-field of UDA \cite{hoffman2018cycada, xu2020adversarial, tzeng2017adversarial, ganin2016domain, tang2020discriminative}. Adversarial-based UDA can be grouped into generative and non-generative categories. The methods of the generative category attempt at generating samples to aid the final classification of the target samples. For example, CyCADA \cite{hoffman2018cycada} adapts source samples to appear as if they are from the target domain, and then trains a category classifier on these adapted source images with their true labels to classify the target data. Similarly, DM-ADA trains the generated auxiliary images that are source-like and the category classifier together from the embeddings of the source and the target domains. The non-generative methods attempt to achieve domain confusion, which usually requires a generator/encoder and a domain label discriminator to engage in a mini-max game. The discriminator attempts at recognizing the domain label of a given sample, and the generator/encoder attempts at masking the source images to be target-like, or vice versa. For example, ADDA \cite{tzeng2017adversarial} makes the generator/encoder on the target samples train against a domain label discriminator, and the goal is to obtain a target generator/encoder that can successfully mask the target samples as if they were from the source domain. Consequently, a category classifier trained on the source samples and their true labels can be used to classify these encoded target samples. RSDA's \cite{gu2020spherical} idea on how to achieve UDA is similar to the vanilla non-generative idea with the mini-max game, with a twist that they define the neural networks in the spherical feature space. Our E-ADDA is in the non-generative category.

\subsection{Out-Of-Distribution Detection}
A lot of attention has been paid to detecting abnormal samples so that they can be intercepted before being sent to a neural network. Specifically, \cite{hendrycks2016baseline}, \cite{lee2018simple}, and \cite{liang2017enhancing} are three state-of-the-art approaches to detect out-of-distribution samples. Liang et al. \cite{liang2017enhancing} observe that fabricating small perturbations into samples as well as using temperature scaling can separate the softmax scores of in-distribution and out-of-distribution samples. Lee et al. \cite{lee2018simple} use Mahalanobis distance to separate in-distribution samples from out-of-distribution ones. 

Lee et al. \cite{lee2018simple} provide a comparison of the three approaches and the performances of the three approaches are indicated in Table \ref{table:ood_algs}, from which we observe that the Mahalanobis distance based approach outperforms Softmax Probability \cite{hendrycks2016baseline} and ODIN \cite{liang2017enhancing}. Therefore, in the rest of the paper, we use the Mahalanobis distance-based approach for the out-of-distribution detection part of our solution. For details, see Section 3.3. 

\begin{table}[h]
    \centering
    \begin{tabular}{ ccccc } 
    \hline
    & Softmax Probability     & Mahalanobis   & ODIN\\
    \hline
    Acc. & 85.06\%      & \textbf{95.75\%}      & 91.08\% \\
       
    \hline
    \\
    \end{tabular}
    \caption{The performances of the three state-of-the-art out-of-distribution detection algorithms. The metric is accuracy. The performances are obtained when training ResNet on CIFAR-10 and SVHN samples are used as out-of-distribution samples. }
   \label{table:ood_algs}
\end{table}

There have not been enough works on incorporating out-of-distribution detection with transfer learning or domain adaptation. Perera et al. \cite{Perera_2019_CVPR} use an out-of-distribution dataset to improve the performance of a classifier on in-distribution samples, which is the only work that intends to combine the two knowledge fields. Our work, E-ADDA, is one of the first approaches that use out-of-distribution to improve the performance of (unsupervised) domain adaptation.

\section{Enforced Adversarial Discriminative Domain Adaptation (E-ADDA)}

\begin{figure}
\centering
\includegraphics[width=0.5\textwidth]{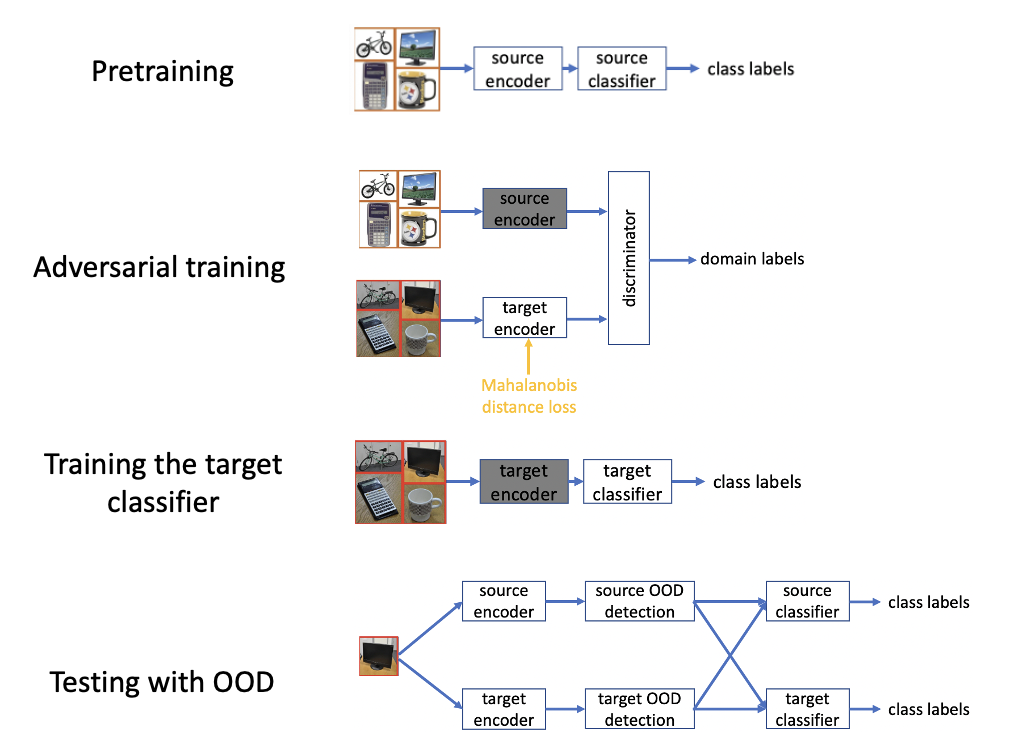}
\caption{The flowchart of E-ADDA: the pretraining, adversarial training, target category classifier training and testing phases. In the pretraining phase, the source encoder $E_s$ and the source category classifier $F_s$ are trained end-to-end using the source samples and their labels. In the adversarial training phase, we freeze the source encoder $E_s$ and train the target encoder $E_t$ and the discriminator $D$ adversarially by engaging them in a mini-max game. To train $E_t$, in addition to the adversarial loss, we incorporate the new Mahalanobis distance loss defined in Equation \ref{eq:mahalanobis_loss}. To train the target category classifier $F_t$, we freeze the adversarially trained $E_t$ and train $F_t$ using its outputs on the target samples. Note that $F_t$ is trained using the pseudo-labels of the target domain samples. During the testing phase, each sample $x$ (in the testing set of) the target domain, $E_s(x)$ and $E_t(x)$ are calculated to determine if the domain confusion is successful. If the domain confusion is not successful, $E_t(x)$ is sent to the target category classifier $F_t$ instead of $F_s$.}
\label{fig:e_adda_flowchart}
\end{figure}

\subsection{Settings of Unsupervised Domain Adaptation}
In UDA, the source samples and their labels are available. The source data is represented as $\boldsymbol{\mathcal{X}}_s = \{ (x^i_s, y^i_s) \}^{N_s}_{i=1}$. Only the samples of the target domain are available; their labels are not available. The target data is represented as $\boldsymbol{\mathcal{X}}_t = \{ (x^i_t) \}^{N_t}_{i=1}$. ${N_s}$ and ${N_t}$ represent the sizes of the sets of the source and target domains, respectively.

\subsection{Adversarial Training with the New Mahalanobis Distance Loss}

Because E-ADDA is based on ADDA \cite{tzeng2017adversarial}, we briefly recap the architecture of ADDA. In ADDA, there exist four neural networks: the source encoder/generator $E_s$, the source category classifier $F_s$, the target encoder/generator $E_t$, and the domain label discriminator $D$. $E_s$ and $F_s$ are trained end-to-end using the true labels of the source samples. Then, with $E_s$ as an input, $E_t$ and $D$ engage in a mini-max game in which $E_t$ tries to mask the target samples to appear as if they are (source samples that are) encoded by $E_s$, and $D$ tries to spot its trick and recover the true origin (source or target) of an encoded sample. Thus, domain confusion is achieved as $E_t$ is able to mask the target data to appear source-like, and $F_s$ is then able to classify them with satisfactory performance.

In E-ADDA, we keep the four neural networks and the framework of ADDA unchanged. One key thing added is the new Mahalanobis distance loss function to train $E_t$ to further enhance/enforce domain confusion. In the following equations, we define the loss function for $E_s$, $E_t$, $F_s$, and $D$.

The source category classifier $F_s$'s loss is the standard supervised loss. It is noted that $E_s$ and $F_s$ are trained jointly, which is achieved by Equation \ref{eq:loss_E_s_R_s}.
\begin{equation}
\label{eq:loss_E_s_R_s}
\begin{aligned}
\underset{E_s, F_s}{\min} \: \mathcal{L}_{F_s}(\boldsymbol{X}_s, Y_s)
& = \mathds{E}_{(x_s, y_s) \sim (\boldsymbol{X}_s, Y_s)} \\
& - \sum^{K}_{k=1} \log F_s(E_s(x_s)) \mathbb{1}(k, y_s)
\end{aligned}
\end{equation}

The domain label discriminator $D$ is also trained using the standard supervised loss using $E_s$ and $E_t$ as well as the domain information of samples in the source and target domains, as in Equation \ref{eq:loss_D}.

\begin{equation}
\label{eq:loss_D}
\begin{aligned}
\mathcal{L}_D(\boldsymbol{X}_s, \boldsymbol{X}_t, E_s, E_t) = & -\mathds{E}_{x_s \sim \boldsymbol{X}_s} [\log D(E_s(x_s)) ] \\
& -\mathds{E}_{x_t \sim \boldsymbol{X}_t} [\log(1 -D(E_t(x_t)))]
\end{aligned}
\end{equation}

In this paragraph, we describe the adversarial training loss and the new Mahalanobis distance loss for $E_t$, as in Equation \ref{eq:loss_E_t}. Note that we call this Mahalanobis loss ``new" to distinguish it from the Mahalanobis distance loss defined by Wen et al. \cite{wen2022rotated}, who take the true values and predicted values as input. In other words, the previously defined Mahalanobis distance loss \cite{wen2022rotated} minimizes the distance between a predicted value and the distribution of the set of true values. Our \textit{new} Mahalanobis distance loss, on the other hand, minimizes the distance between two distributions.

\begin{equation}
\label{eq:loss_E_t}
\begin{aligned}
\mathcal{L}_{E_t}(\mathcal{X}_s, \mathcal{X}_t, D) = &-\underset{d \in \{s, t\} }{\sum} \mathds{E}_{x_d \sim \boldsymbol{X}_d}  [\frac{1}{2}\log D(E_d(x_d)) ] \\
&+ [\frac{1}{2}\log(1 - D(E_d(x_d))) ] + \theta_M \mathcal{L}_M
\end{aligned}
\end{equation}

How do we define $\mathcal{L}_M$, the new Mahalanobis distance loss that minimizes the distance between two distributions? To define it, we consider the domain confusion task to be achieved. We want to train $E_t$ and $D$ adversarially so that $E_t$ encodes the target samples such that $D$ thinks these encoded samples were source samples encoded by $E_s$. Therefore, to further maximize domain confusion, we define $\mathcal{L}_M$ as Equation \ref{eq:mahalanobis_loss}. $\hat{\mu}_s$ is the empirical mean of all source samples encoded by $E_s$ defined as Equation \ref{eq:emp_mean}, and $\hat{\Sigma}$ is the empirical covariance defined as Equation \ref{eq:emp_covar}.
\begin{equation}
\label{eq:mahalanobis_loss}
\begin{aligned}
\mathcal{L}_{M} = \sum (E_t(x_t)-\hat{\mu}_s)^{\top} \hat{\Sigma_s}^{-1}(E_t(x_t)-\hat{\mu}_s)
\end{aligned}
\end{equation}

\begin{equation}
\label{eq:emp_mean}
\begin{aligned}
\hat{\mu}_s = \frac{1}{N_s} \sum E_s(x_s), x_s \in \boldsymbol{X}_s
\end{aligned}
\end{equation}

\begin{equation}
\label{eq:emp_covar}
\begin{aligned}
\Sigma_s = \frac{1}{N_s} \sum (E_s(x_s) - \hat{\mu}_s) (E_s(x_s) - \hat{\mu}_s)^{\top}
\end{aligned}
\end{equation}

\subsection{(Traditional) Mahalanobis Distance Based Out-of-Distribution Detection Subroutines}

To safeguard the scenario in which domain confusion still somehow fails despite our best effort with the new Mahalanobis loss, we add two OOD detection subroutines to catch a (target) sample if the adversarial training fails to allow $E_t$ to mask it as if it was a source sample encoded by $E_s$. If this happens, we send this target sample to the target category classifier, instead of the source category classifier, for final classification. The target classifier is trained using the target training samples and their pseudo-labels.

To determine if a sample $x$ is still within the distribution of the target domain or if it is successfully encoded to look like its origin is the source domain, we measure the Mahalanobis distance between $E_s(x)$ and the set of $E_s(x_s)$, $\forall x_s \in \boldsymbol{X}_s$, as well as the Mahalanobis distance between $E_t(x)$ and the set of $E_t(x_t)$, $\forall x_t \in \boldsymbol{X}_t$. Note that this Mahalanobis distance has nothing to do with the previously defined $\mathcal{L}_M$; in the current case of OOD detection subroutines, we apply the ``traditional" definition of the Mahalanobis distance, which measures how far a value diverges from a distribution. To get the parameters that the calculation of the traditional Mahalanobis distance requires, we need the empirical mean and the empirical covariance of the distribution. For the source distribution, we have already defined the source empirical mean $\hat{\mu}_s$ in Equation \ref{eq:emp_mean} and the source empirical covariance $\hat{\Sigma}_s$ in Equation \ref{eq:emp_covar}. Similarly, we define the target empirical mean $\hat{\mu}_t$ and the target empirical covariance $\hat{\Sigma}_t$ in Equations \ref{eq:emp_mean_tgt} and \ref{eq:emp_covar_tgt}.

\begin{equation}
\label{eq:emp_mean_tgt}
\begin{aligned}
\hat{\mu}_t = \frac{1}{N_t} \sum E_t(x_t), x_t \in \boldsymbol{X}_t
\end{aligned}
\end{equation}

\begin{equation}
\label{eq:emp_covar_tgt}
\begin{aligned}
\Sigma_t = \frac{1}{N_t} \sum (E_t(x_t) - \hat{\mu}_t) (E_t(x_t) - \hat{\mu}_t)^{\top}
\end{aligned}
\end{equation}

The traditional Mahalanobis distance between $x$ and a distribution is defined using an empirical mean $\hat{\mu}$ and empirical covariance $\hat{\Sigma}$ that describe the distribution, as defined in Equation \ref{eq:mahalanobis}, in which $E$ can be either $E_s$ or $E_t$, depending on if this is the source or the target distribution that we are talking about.

\begin{equation}
    \begin{aligned}
    \hat{M}(x) = (E(x)-\hat{\mu})^\top \hat{\Sigma}^{-1} (E(x)-\hat{\mu})
    \end{aligned}
    \label{eq:mahalanobis}
\end{equation}

With the aforementioned information, we create two OOD subroutines. The first one checks the traditional Mahalanobis distance between $E_s(x)$ and the distribution of $E_s(x_s)$, $\forall x_s \in \boldsymbol{X}_s$, The second one checks the Mahalanobis distance between $E_t(x)$ and the distribution of $E_t(x_t)$, $\forall x_t \in \boldsymbol{X}_t$. If the Mahalanobis distance score between $E_s(x)$ and the distribution of $E_s(x_s)$ is smaller than an empirically determined $\lambda_s$, $\forall x_s \in \boldsymbol{X}_s$ then it is considered within the distribution of the source domain. If the traditional Mahalanobis distance score between $E_t(x)$ and the distribution of $E_t(x_t)$, $\forall x_t \in \boldsymbol{X}_t$ is smaller than an empirically determined $\lambda_t$, then it is considered within the distribution of the target domain. 



\section{Evaluation}
\label{sec:evaluation}
\subsection{Overview}
We first evaluate E-ADDA on a domain adaptation task on an acoustic modality: we domain-adapt from a dataset consisting of emotional utterances to a dataset that contains audio samples of speech in which, sometimes, the speakers are in a verbal conflict (we map the anger emotion to conflict and other emotions to non-conflict). Then, to demonstrate that E-ADDA not only works on domain-adapting from the domain of emotions to the domain of conflict speech, but also in other fields such as computer vision, we compare E-ADDA against various other state-of-the-art deep domain adaptation algorithms on standard datasets and tasks of UDA in the field of computer vision such as Office-31 and Office-Home.

\subsection{The Domain Adaptation Task on the Acoustic Modality}
\subsubsection{The Source Dataset}
In the following paragraphs we describe our source dataset in the domain adaptation task on the acoustic modality. The EMOTION dataset \textbf{contains the all samples from the following 5 public datasets}: RAVDESS \cite{livingstone2018ryerson}, CREMA-D \cite{cao2014crema}, EMA \cite{lee2005articulatory},  TESS \cite{dupuis2010toronto}, and SAVEE \cite{HaqJackson_MachineAudition10}. In addition we extend these 5 datatsets with samples that are distorted to account for environmental conditions by artificially adding environmental distortions into the clean samples from the original five datasets. The reverberation effect is described by the combination of the decay factor,  diffusion, and wet/dry ratio. EMOTION consists of training and testing sets. In the training set, there are 8,816 samples in the anger class, 8,786 samples in the happiness class, 7,742 samples in the neutral class, 8,811 samples in the sadness class, and 5,761 samples in the disgust/fear class. In the testing set, there are 1,942 samples in the anger class, 1,966 samples in the happiness class, 1,696 samples in the neutral class, 1,947 samples in the sadness class, and 1,292 samples in the fear/disgust class. 

\subsubsection{The Target Dataset}
Our target dataset, CONFLICT, is the dataset where we want to apply E-ADDA solution so that the source classifier trained on EMOTION can be re-purposed. It is collected from real home environments in which 19 couples talk (collected with approved IRB) about topics that they previously disagree on and have their conversation recorded. In total, there are 3027 training samples and 1009 testing samples.

\subsubsection{Comparison with State-of-the-Art Baselines}

In this experiment, shown in Table \ref{table:eval_baselines}, we compare E-ADDA with the scenario in which no domain adaptation or transfer learning is used (No DA/TL) and three baselines: direct training (directly training the model on the data from the target dataset), two selected state-of-the-art approaches, ADDA and ADDA with CORAL loss. Our solution, E-ADDA, outperforms various state-of-the-art domain adaptation/transfer learning algorithms on the acoustic modality in the field of domain-adapting/transfer-learning from angry voices to speeches of verbal conflict by up to 29.8\% improvement in f1 scores.

Each of the audio samples on which we test the situation in which no DA/TL is used, the three baselines, and E-ADDA contains environmental distortions and/or overlapped speech. The usage of CORAL loss (in Deep CORAL) and ADDA has garnered a lot of interest in the field of DA/TL; in this paragraph, we briefly describe these two approaches. CORAL loss proposes that the domain shift can be mitigated by using linear transformations to align the second-order statistics of the two domains. ADDA proposes to encode the target samples to the feature space of the source and have a domain discriminator that tries to distinguish encoded target samples from source samples. ADDA and ADDA with CORAL loss achieve in f1 scores of 38.29\% and 63.28\% respectively.

\begin{table}[h]
    \centering
    \begin{tabular}{ cccc } 
    \hline
                & Env. Distortion     & F1\\
    \hline
    
    ADDA            & \checkmark    & 38.29\%  \\ 
    ADDA + CORAL     & \checkmark    & 63.28\%  \\ 
    No TL/DA     & \checkmark    & 77.25\%  \\ 
    Trained on target & \checkmark & 85.82\% \\
    E-ADDA    & \checkmark & \textbf{93.10\%}\\ 

    \hline
    \\
    \end{tabular}
    \caption{The performance of four baselines against E-ADDA on data that has overlapped speech and environmental distortions.}

    \label{table:eval_baselines}
\end{table}

As shown in Table \ref{table:eval_baselines}, No TL/DA's performance is 77.25\%, a value that is higher than the state-of-the-art solutions ADDA and ADDA with CORAL loss. No TL/DA stands for that we directly apply the source classifier on the target samples. In the case of domain-adapting from a classifier of emotions to conflict detection, no TL/DA suggests that we directly apply the mood classifier on the conflict samples and the performance is calculated based on that anger denotes conflict while other emotions denote no conflict. ADDA performance was 38.29\%, which is lower than the no TL/DA by 38.96\%. ADDA with CORAL loss achieved significantly higher performance, 63.28\%. Since with have more than 7000 samples in the target dataset, we also directly train a classifier using only the target sample and yield an f1 score of 85.82\%, which is higher than ADDA by 47.53\% and ADDA combined with CORAL loss by 22.54\%. Still, it is 7.28\% lower than E-ADDA's performance. Our E-ADDA results in an improvement over ADDA with CORAL loss by 29.82\%.

It is worth noting that the source and target domains in this setting are distribution-wise distant because they are not even from the same class (the source domain is about people's emotions and the target domain is about verbal conflict). Therefore, the task should be more appropriately called unsupervised transfer learning instead of unsupervised domain adaptation. We present this task as part of our evaluation to test if E-ADDA can really enforce domain confusion, safeguard catching samples on which domain confusion fails, and send these samples to their respective category classifiers. The ADDA architecture only yields an f1 score of 38.2\%, suggesting that ADDA's basic mechanism of domain confusion fails. However, with the {new} Mahalanobis distance loss and the OOD detection subroutine on top of the same architecture, E-ADDA is able to achieve an f1 score of 93.1\%. This indicates that the {new} Mahalanobis distance loss is very effective at enforcing, on top of the adversarial training, domain confusion. In addition, it suggests the necessity of the OOD detection subroutine to send samples on which domain confusion fails to their respective category classifiers.

\begin{table*}[h]
    \centering
    \begin{tabular}{ ccccccc|c } 
    
    \hline
    Algorithm   &  A$\rightarrow$W &  A$\rightarrow$D &  D$\rightarrow$W &  D$\rightarrow$A &  W$\rightarrow$A &  W$\rightarrow$D  & Avg    \\
    \hline
    ResNet-50 \cite{he2016deep}    &   68.4\%&            68.9\%&             96.7\%&         62.5\%&         60.7\%&   99.3\%&           76.1\%\\
    DANN \cite{ganin2015unsupervised}         &   82.0\%&            79.7\%&             96.9\%&         68.2\%&         67.4\%&   99.1\%&           82.2\%\\
    MSTN \cite{xie2018learning}        &   91.3\%&            90.4\%&             98.9\%&         72.7\%&         65.6\%&   \textbf{100\%}&   86.5\%\\
    CDAN+E \cite{long2018conditional}      &   94.1\%&            92.9\%&             98.6\%&         71.0\%&         69.3\%&   \textbf{100\%}&   87.7\%\\
    DMRL \cite{wu2020dual}         &   90.8\%&            93.4\%&             99.0\%&         73.0\%&         71.2\%&   \textbf{100\%}&   87.9\%\\
    SymNets \cite{zhang2019domain}      &   90.8\%&            93.9\%&             98.8\%&         74.6\%&         72.5\%&   \textbf{100\%}&   88.4\%\\
    GSDA \cite{hu2020unsupervised}         &   95.7\%&            94.8\%&             99.1\%&         73.5\%&         74.9\%&   \textbf{100\%}&   89.7\%\\
    CAN \cite{kang2019contrastive}         &   94.5\%&            95.0\%&             99.1\%&         78.0\%&         77.0\%&   99.8\%&           90.6\%\\
    SRDC \cite{tang2020unsupervised}        &   95.7\%&            95.8\%&    99.2\%&         76.7\%&         77.1\%&   \textbf{100\%}&   90.8\%\\
    RSDA-MSTN \cite{gu2020spherical}    &   \textbf{96.1\%}&   95.8\%&    99.3\%&         77.4\%&         78.9\%&   \textbf{100\%}&   91.1\%\\
    \hline
    E-ADDA    &   95.4\%&            \textbf{96.2\%}&             \textbf{100\%} &   \textbf{95.3\%}&   \textbf{90.9\%} &   \textbf{100\%}&   \textbf{95.3\%}\\

    \hline
    \\
    \end{tabular}
    \caption{The results on the domain adaptation tasks among the three domains in the dataset Office-31. The metric is accuracy.}

    \label{table:eval_office_31}
\end{table*}

\begin{table*}[h]
    \centering
    \scalebox{0.6}{
    \begin{tabular}{ ccccccccccccc|c } 
    \hline
    Algorithm   & Pr$\rightarrow$Ar & Ar$\rightarrow$Pr & Cl$\rightarrow$Ar & Ar$\rightarrow$Cl & Rw$\rightarrow$Ar & Ar$\rightarrow$Rw & Pr$\rightarrow$Cl & Cl$\rightarrow$Pr& Rw $\rightarrow$ Pr & Pr$\rightarrow$Rw & Rw$\rightarrow$Cl & Cl$\rightarrow$Rw  & Avg    \\
    \hline
    ResNet-50 \cite{he2016deep} & 38.5\% & 50\% & 37.4\% &  34.9\% &  53.9\% &  58\% &  31.2\% &  41.9\% & 59.9\% &  60.4\% &  41.2\% &  46.2\% & 46.1\% \\
    
    DANN \cite{ganin2015unsupervised}        &  41.6\% &  59.3\% &  47.0\% &  45.6\% &  63.2\% &  70.1\% &  43.7\% &  58.5\% &  76.8\% &  68.5\% &  51.8\% &  60.9\% & 57.6\%\\
    CDAN \cite{long2018conditional}       &  55.6\% &  69.3\% &  54.4\% &  49.0\% &  68.4\% &  74.5\% &  48.3\% &  66.0\% &  80.5\% &  75.9\% &  55.4\% &  68.4\% & 63.8\%\\
    MSTN \cite{xie2018learning}       &  61.4\% &  70.3\% &  60.4\% &  49.8\% &  70.9\% &  76.3\% &  48.9\% &  68.5\% &  81.1\% &  75.7\% &  55.0\% &  69.6\% & 65.7\%\\
    SymNets \cite{zhang2019domain}     &  63.6\% &  72.9\% &  64.2\% &  47.7\% &  73.8\% &  78.5\% &  47.6\% &  71.3\% &  82.6\% &  79.4\% &  50.8\% &  74.2\% & 67.2\% \\
    GSDA \cite{hu2020unsupervised}         &  65.0\% &  76.1\% &  65.4\% &  61.3\% &  72.2\% &  79.4\% &  53.2\% &  73.3\% &  83.1\% &  80.0\% &  60.6\% &  74.3\% & 70.3\%\\
    GVB-GD \cite{cui2020gradually}      &  65.2\% &  74.7\% &  64.6\% &  57.0\% &  74.6\% &  79.8\% &  55.1\% &  74.1\% &  84.3\% &  81.0\% &  59.7\% &  74.6\% & 70.4\%\\
    RSDA-MSTN \cite{gu2020spherical}   &  67.9\% &  77.7\% &  66.4\% &  53.2\% &  75.8\% &  \textbf{81.3\%} &  53.0\% &  74.0\% &  85.4\% &  \textbf{82.0\%} &  57.8\% &  76.5\% & 70.9\%\\
    SRDC \cite{tang2020unsupervised}        &  \textbf{68.7\%} &  76.3\% &  \textbf{69.5\%} &  52.3\% &  \textbf{76.3\%} &  81.0\% &  53.8\% &  76.2\% &  85.0\% &  81.7\% &  57.1\% &  \textbf{78.0\%} & 71.3\%\\

    \hline
    E-ADDA   & 66.8\% & \textbf{78.6\%} & 59.6\% & \textbf{61.0\%} & 67.7\% & 79.7\% & \textbf{64.9\%} & \textbf{79.8\%} & \textbf{85.8\%} & 79.2\% & \textbf{64.9\%} & 70.4\% & \textbf{71.5\%} \\
    \hline
    \\
    \end{tabular}
    }
    \caption{The results on the domain adaptation tasks among the four domains in the dataset Office-Home. The metric is accuracy.}

    \label{table:eval_office_home}
\end{table*}

\begin{table}[h!]
\begin{center}
\scalebox{0.7}{
    \begin{tabular}{ ccc } 
    \hline
    Algorithm & MNIST $\rightarrow$ USPS & SVHN $\rightarrow$ MNIST \\
    \hline
    Source only         & 75.2\%        & 60.1\% \\
    Gradient Reversal \cite{ganin2015unsupervised}   & 77.1\%        & 73.9\% \\
    Domain Confusion \cite{tzeng2015simultaneous}   & 79.1\%        & 68.1\% \\
    CoDAN \cite{liu2016coupled}               & 91.2\%        & did not converge \\
    ADDA \cite{tzeng2017adversarial}                & 89.4\%        & 76.0\% \\
    Associative \cite{haeusser2017associative}        & 94.1\%        & 93.6\% \\
    DANN \cite{ganin2016domain}                & 60.8\%        & 76.3\% \\
    Deep Coral \cite{sun2016deep}          & 69.5\%        & 76.3\% \\
    VADA \cite{shu2018dirt}                & 90.6\%        & 92.6\% \\
    \hline
    E-ADDA         & \textbf{95.4\%}        & \textbf{95.4\%} \\
    \hline
    \\
    \\
    \end{tabular}
    }
    \caption{We compare our technique, E-ADDA, with nine other state-of-the-art deep domain adaptation techniques on two tasks (the performance is measured in accuracy, per the evaluation standard of the computer vision community).}
    
    \label{tab:eval_compare_against_baselines}
\end{center}
\end{table}

\begin{table}[h!]
\begin{center}
\scalebox{0.7}{
    \begin{tabular}{ cc } 
    \hline
    Algorithm & STL-10 $\rightarrow$ CIFAR-10 \\
    \hline
    DRCN \cite{ghifary2016deep}                 & 58.6\%        \\
    SE \cite{french2017self}       & 64.2\%        \\
    Source only                                 & 63.6\%        \\
    VADA \cite{shu2018dirt}                     & 75.3\%        \\
    Co-DA \cite{kumar2018co}                    & 76.4\%        \\
    DTA \cite{kumar2018co}                      & 72.8\%        \\
    ET                                          & \textbf{86.1\%}        \\
    \hline
    \\
    \\
    \end{tabular}
    }
    \caption{We compare our technique, E-ADDA, with five other state-of-the-art deep domain adaptation techniques on the domain adaptation task to domain-adapt from STL-10 to CIFAR-10 (the performance is measured in accuracy, per the evaluation standard of the computer vision community).}
    
    \label{tab:eval_compare_against_baselines_cifar_stl}
\end{center}
\end{table}


\subsection{Domain Adaptation Tasks on Images}
In this section, we discuss the performance of E-ADDA against state-of-the-art baselines on popular benchmarks for UDA such as Office-31 and Office-Home. Then, to show that E-ADDA also achieves state-of-the-art performance on simpler domain adaptation tasks such as MNIST $\rightarrow$ USPS, SVHN $\rightarrow$ MNIST, as well as CIFAR-10 $\rightarrow$ STL-10, we also compare E-ADDA's performance against state-of-the-art baselines on these UDA tasks.

\subsubsection{Office-31}

In Table \ref{table:eval_office_31}, we compare our E-ADDA against ResNet-50 \cite{he2016deep} and nine other state-of-the-art domain adaptation algorithms using the dataset Office-31. Office-31 contains three subdomains: Amazon (A), Webcam (W), and Dslr (D). Each domain contains 31 classes of everyday office objects such as rulers or projectors. There are 4,110 images in total in Office-31. Across the three domains, six domain adaptation tasks can be formed, as shown in Table \ref{table:eval_office_31}. The performance of each algorithm is measured in the accuracy that is the percentage of samples that are correctly classified by the algorithm out of all the samples in the testing set. 

On the six domain adaptation tasks, we have achieved state-of-the-art performance on five of them, except for the task of A $\rightarrow$ W, where RSDA-MSTN \cite{gu2020spherical} outperforms E-ADDA by 0.7\%. RSDA-MSTN \cite{gu2020spherical} proposes to redefine the feature space as a spherical feature space and create a spherical classifier and discriminator, creating a pseudo-label loss in this spherical feature space. However, it fails to deal with the situation in which the pseudo-labels are not very accurate and the pseudo-label loss is very large. E-ADDA does not have that problem. 

It is worth noting that RSDA-MSTN is a non-generative adversarial algorithm whose superiority comes from the fact that the adversarial training is defined in the spherical feature space. As a non-generative adversarial method, RSDA-MSTN is a perfect candidate to compare E-ADDA against. On tasks with large domain shifts, such as W $\rightarrow$ A and D $\rightarrow$ A, we outperform RSDA-MSTN by 17.9\% and 12\%, a very large improvement. This suggests E-ADDA is better at achieving domain confusion on UDA tasks whose source and target domains are more distributionally distant while other algorithms that aim at domain confusion fail to achieve a performance that is as high.

\subsubsection{Office-Home}

In Table \ref{table:eval_office_home}, we compare E-ADDA against ResNet-50 and eight other state-of-the-art domain adaptation algorithms on Office-Home. Office-Home has four subdomains: Product (Pr), Art (Ar), Clipart (Cl), and Real World (Rw). There are 15,500 images in Office-Home, each of which is of a typical object that can be found in an office or home, such as flowers. Twelve domain adaptation algorithms can be formed based on the four subdomains. The performance of each algorithm is measured in the accuracy that is the percentage of samples that are correctly classified by the algorithm out of all the samples in the testing set.

Out of the twelve domain adaptation tasks, we outperform the next best-performing algorithm on six of them. On the task of Ar $\rightarrow$ Rw, the state-of-the-art, RSDA-MSTN, outperforms us by 1.6\%. Again, RSDA fails to deal with the situation in which the pseudo-labels are not very accurate and the pseudo-label loss is very large. On the task Pr $\rightarrow$ Ar, the state-of-the-art, SRDC, outperforms us by 1.9\%. SRDC proposes to alleviate the risk of damaging the intrinsic domain discrimination resulting from finding domain-aligned features. However, the proposition to minimize the KL divergence between the distribution of predictive labels and the distribution of auxiliary labels is a rather naive approach, as the authors fail to compare their algorithm with other measurements to minimize the Jensen–Shannon divergence.

Once again, we have observed that E-ADDA is better at achieving domain confusion than the other non-generative adversarial method, RSDA-MSTN, when domain shifts are large. For example, on the task Ar $\rightarrow$ Cl, we outperform the second-best-performing algorithm RSDA-MSTN by 7.7\%. This suggests that, when domain shifts are large, or when domain confusion is harder, E-ADDA can still achieve good domain confusion results, while the state-of-the-art non-generative methods cannot.

\subsubsection{MNIST $\rightarrow$ USPS and SVHN $\rightarrow$ MNIST}

MNIST, USPS, and SVHN datasets all have ten classes of hand-written digits. The first task, which is to domain-adapt from MNIST to USPS, is considered easier while the second task, which is to domain-adapt from SVHN to MNIST, is considered more challenging. This claim is supported by the observation that, in Table \ref{tab:eval_compare_against_baselines}, five out of the state-of-the-art domain adaptation algorithms and the baseline of directly using source classifier on the target dataset results in lower performance of the second task compared to the first task. E-ADDA achieve an accuracy of 95.41\% on the first task and 95.43\% on the second. On the first task, it outperforms the best-performing state-of-the-art baseline, Associative, by 1.31\%. On the second task, it outperforms the best-performing state-of-the-art baseline by 2.83\%. We present our evaluation results on MNIST $\rightarrow$ USPS and SVHN $\rightarrow$ MNIST to demonstrate that E-ADDA can also achieve state-of-the-art performance on simpler UDA tasks.

\subsubsection{STL-10 $\rightarrow$ CIFAR-10}

In this section, we further investigate if the E-ADDA can outperform state-of-the-art baselines on a more complicated vision task that is not digits. Therefore, we transfer learn from STL-10 to CIFAR-10. Both CIFAR-10 and STL-10 are image datasets that contain 10 classes. 
We outperform all the other five state-of-the-art deep domain adaptation baselines and outperform the second best-performing algorithm, Co-DA, by 9.7\%. The source classifier that we use to train is ResNet-50 \cite{he2016deep}. We yield the highest performance of an accuracy score of 86.1\% after we inject the E-ADDA Cell after the fifth layer.

Note that, compared to the previous section, we choose a different set of baselines to fully evaluate our solution, the E-ADDA, against as many baselines as possible. The task to domain-adapt from STL-10 to CIFAR-10, which is more complex than domain-adapt among MNIST, SVHN, and USPS, as these three datasets contain only digits. On the contrary, CIFAR-10 and STL-10 contains images such as the automobile and dog classes. Again, we present our evaluation results on STL-10 $\rightarrow$ CIFAR-10 to demonstrate that E-ADDA can also achieve state-of-the-art performance on simpler UDA tasks.

\section{Discussion}
In Section \ref{sec:evaluation}, we discussed E-ADDA's performance on two sets of tasks: the first is for verbal conflict detection, and the second is for computer vision. Both sets of tasks are widely used in smart computing. For example, verbal conflict detection is part of smart health applications that monitor mental health, and computer vision is a major component of many smart computing applications such as autonomous vehicles and smart health image analysis. 

\section{Conclusion}
In smart computing, researchers want to perform classification or prediction on a set of data. However, directly training on these data points is not always easy, because the label information of these data is not readily available. However, if there exists a relevant (albeit different) dataset with available label information, then UDA can be performed to use the label information in a dataset (the source domain) to perform better classification in a different, unlabeled dataset (target domain).

We have discovered that there exists room for improvement on the existing non-generative adversarial UDA algorithms that attempt to achieve domain confusion. The challenge lies in the observation that these algorithms do not explicitly minimize the distance between the distribution of the masked/encoded target samples and the source samples; instead, they let the adversarial training achieve domain confusion rather implicitly. To address this challenge, we propose E-ADDA that uses a novel variation of the Mahalanobis distance loss to minimize the distribution-wise distance between the masked/encoded target domain samples and the source domain samples. Then, the OOD subroutine further eliminates samples on which the domain confusion is unsuccessful. We have performed extensive evaluations on E-ADDA on two modalities: the acoustic modality and the computer vision modality. On both modalities, we outperform the state-of-the-art algorithms and achieve new state-of-the-art performance.


\bibliography{references}
\bibliographystyle{plain}

\end{document}